\documentclass[letterpaper, 10 pt, conference]{ieeeconf}  

\IEEEoverridecommandlockouts                              

\usepackage{graphics} 
\usepackage{adjustbox}
\usepackage{epsfig} 
\usepackage{mathptmx} 
\usepackage{times} 
\usepackage{amsmath} 
\usepackage{amssymb}  
\usepackage{xcolor}
\usepackage{cite}
\usepackage{booktabs}
\usepackage{caption}

\newcount\Comments  
\newcommand{\xuesu}[1]{{\ifnum\Comments=1\textcolor{green}{[xuesu: #1]}\fi}}
\newcommand{\zifan}[1]{{\ifnum\Comments=1\textcolor{blue}{[zifan: #1]}\fi}}
\newcommand{\commentp}[1]{{\ifnum\Comments=1\textcolor{red}{[Peter: #1]}\fi}}
\newcommand{\remove}[1]{{\ifnum\Comments=1\textcolor{red}{[#1]}\fi}}

\title{\LARGE \bf
Dexterous Legged Locomotion in Confined 3D Spaces with Reinforcement Learning
}

\author{Zifan Xu$^{1}$, Amir Hossain Raj$^{2}$, Xuesu Xiao$^{2}$, and Peter Stone$^{1, 3}$
\thanks{$^{1}$The University of Texas at Austin
        {\tt\small zfxu@utexas.edu, pstone@cs.utexas.edu}. $^{2}$George Mason University {\tt\small \{araj20, xiao\}@gmu.edu}. $^{3}$Sony AI. This work has taken place in the Learning Agents Research
Group (LARG) at UT Austin.  LARG research is supported in part by NSF (FAIN-2019844, NRT-2125858), ONR (N00014-18-2243), ARO (E2061621), Bosch, Lockheed Martin, and UT Austin's Good Systems grand challenge.
Peter Stone serves as the Executive Director of Sony AI America and receives financial compensation for this work.  The terms of this arrangement have been reviewed and approved by the University of Texas at Austin in accordance with its policy on objectivity in research.}
}

\usepackage{fancyhdr}

\begin{document}

\maketitle

\thispagestyle{fancy}


\begin{abstract}
Recent advances of locomotion controllers utilizing deep reinforcement learning (RL) have yielded impressive results in terms of achieving rapid and robust locomotion across challenging terrain, such as rugged rocks, non-rigid ground, and slippery surfaces. However, while these controllers primarily address challenges \emph{underneath} the robot, relatively little research has investigated legged mobility through confined 3D spaces, such as narrow tunnels or irregular voids, which impose \emph{all-around} constraints. The cyclic gait patterns resulted from existing RL-based methods to learn parameterized locomotion skills characterized by motion parameters, such as velocity and body height, may not be adequate to navigate robots through challenging confined 3D spaces, requiring both agile 3D obstacle avoidance and robust legged locomotion. Instead, we propose to learn locomotion skills end-to-end from goal-oriented navigation in confined 3D spaces. To address the inefficiency of tracking distant navigation goals, we introduce a hierarchical locomotion controller that combines a classical planner tasked with planning waypoints to reach a faraway global goal location, and an RL-based policy trained to follow these waypoints by generating low-level motion commands. This approach allows the policy to explore its own locomotion skills within the entire solution space and facilitates smooth transitions between local goals, enabling long-term navigation towards distant goals. In simulation, our hierarchical approach succeeds at navigating through demanding confined 3D environments, outperforming both pure end-to-end learning approaches and parameterized locomotion skills. We further demonstrate the successful real-world deployment of our simulation-trained controller on a real robot.
\end{abstract}

\section{Introduction}
Quadruped robots capitalize on their systems' many Degrees of Freedom (DoFs) and enjoy superior mobility and versatility to locomote through extremely challenging environments, compared to conventional locomotion modalities such as wheeled and tracked systems~\cite{raibert1986legged}. By coordinating all the joint angles on their four limbs, quadruped robots can quickly react to unstructured environments, such as rugged rocks, non-rigid ground, and slippery surfaces, and maintain a stable body pose while moving forward. 

However, such advanced mobility and versatility comes at a price: in order to coordinate the many DoFs to efficiently maintain torso stability and progress forward, sophisticated planning and control algorithms are necessary to run in real time and react quickly to any environmental constraints and disturbances. Classical gait-based controllers, introduced decades ago~\cite{owaki2017quadruped, sprowitz2013towards, lewis2002gait}, are inspired by the quadruped robots' biological counterparts, e.g., dogs and horses, and are able to move them through moderately challenging terrain. 
Recently, with the advances in machine learning, robotics researchers have also applied learning methods to enable emergent quadruped locomotion behaviors~\cite{shao2021learning, fu2021minimizing, tan2018sim, yang2020data, pan2020zero}. These learning-based methods can often produce better locomotion performance compared to classical hand-crafted approaches, enabling complex locomotion skills~\cite{rudin2022advanced, margolis2022rapid}, safe and agile performance~\cite{yang2022safe, hwangbo2019learning}, or superior robustness against a variety of unstructured environments~\cite{kumar2021rma, agarwal2023legged, yu2021visual}. 

\begin{figure}
    \centering
    \includegraphics[width=0.8\columnwidth]{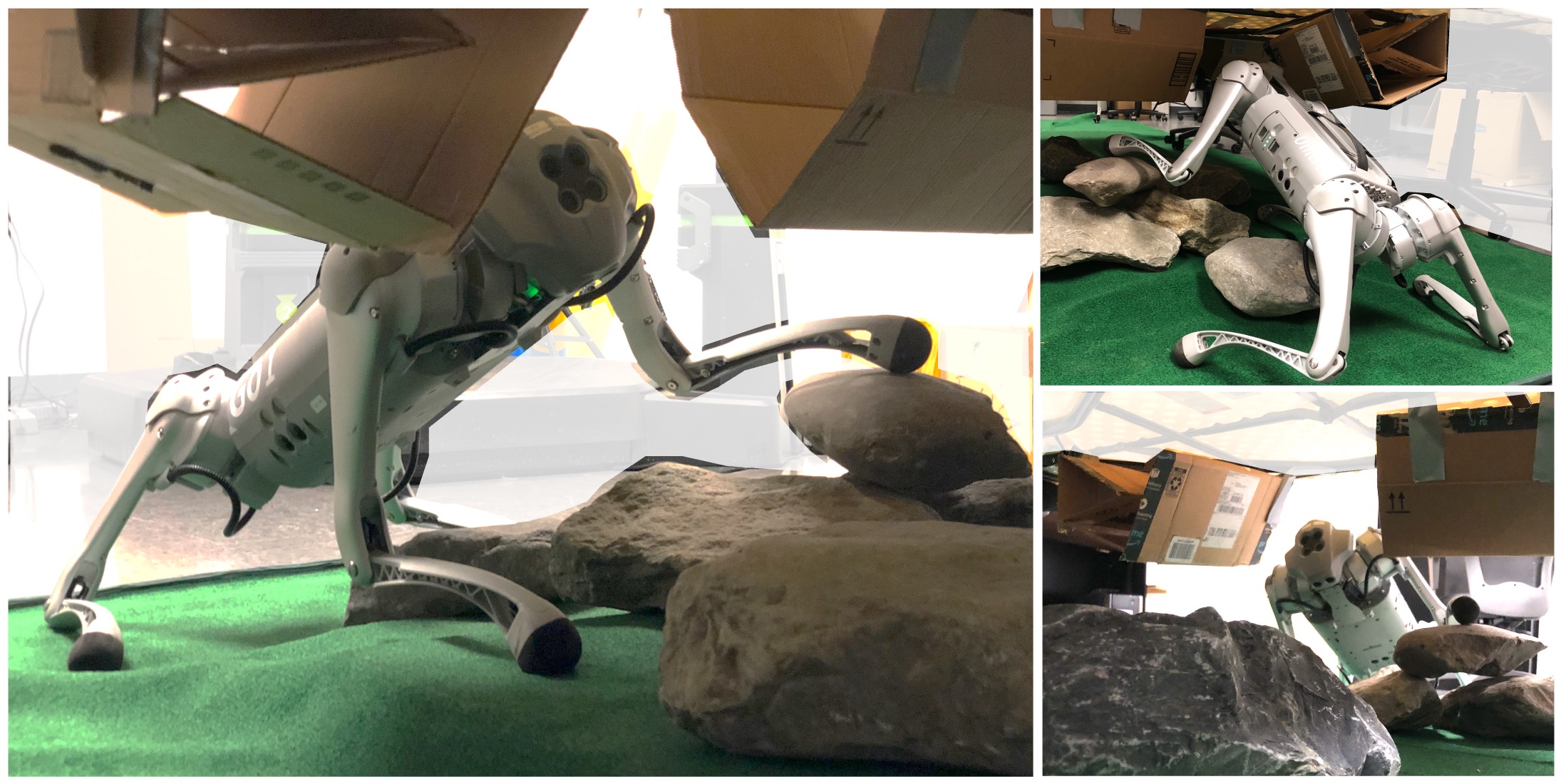}
    \caption{Dexterous quadruped legged locomotion in real-world confined 3D spaces. }
    \label{fig::go1}
\end{figure}

While current quadruped locomotion techniques allow robots to robustly move through different terrain that give rise to challenges from \emph{underneath} the robot, there is limited research on how a hyper-redundant quadruped robot can move through confined 3D environments which impose \emph{all-around} constraints from 360\textdegree~surroundings (Fig.~\ref{fig::go1}), e.g., through narrow cave networks~\cite{rouvcek2020darpa}, complex industrial plants~\cite{bellicoso2018advances}, and cluttered indoor environments (e.g., underneath a coffee table)~\cite{Matterport3D}, where narrow tunnels or irregular voids are not uncommon. In those spaces, robots not only need to maintain stability due to the underlying terrain and protrusions from the ground, but also precisely maneuver their torso and limbs to squeeze between 3D obstacles from walls and ceilings with a variety of body poses, e.g., large roll and pitch angles with asymmetric, acyclic, and irregular limb movements. In some scenarios, a stable torso may not even be possible to maintain, requiring the robot to lean against obstacles to enable obstacle-aided locomotion~\cite{transeth2008snake}. 

These demanding behaviors are not likely to emerge in transitional parameterized locomotion skills trained by an RL agent tracking motion parameters, such as linear and angular velocities and torso orientations. Instead, we hypothesize that such dexterous locomotion, combined with navigation and manipulation, can only be developed by end-to-end learning of goal-reaching tasks in diverse and confined 3D spaces. In addition, to address the inefficiency of reaching faraway navigation goals, we develop a hierarchical locomotion system, which is composed of a high-level sampling-based planner that plans a feasible body pose as a local goal in front of the robot, and a low-level locomotion policy that is trained with RL to reach this local goal. The sampling-based high-level reactive planner is able to quickly respond to the limited and mostly obstructed robot perception in those confined 3D spaces in a computationally efficient manner to assure real-time execution. The low-level RL policy can robustly move the many limb joints to follow the high-level local poses while considering the all-around constraints not only from the terrain underneath but also 360\textdegree-surrounding irregular obstacles. 
Such a hybrid setup allows the low-level RL policy to fully exploit the hyper-redundant solution space with better sample efficiency compared with pure end-to-end learning using sparse goal-reaching reward. By training on sequences of local goals planned by the high-level controller, the low-level RL policy also enables smooth transitions between local goals, which guides the long-term navigation to distant global goals. 
The hierarchical training pipeline outperforms both pure end-to-end learning and learning of parameterized locomotion skills. Our main contributions are as follows: 
\begin{enumerate}
    \item We introduce a novel hybrid approach that combines long-term path planning from a
classical planner with short-term local goal-reaching via an RL-based locomotion controller.
    \item We provide a procedural environment generation pipeline for generating diverse and challenging confined 3D spaces.
    \item We investigate the limitations of pre-defined parameterized locomotion skills and pure end-to-end learning in traversing 3D confined spaces.
\end{enumerate}

\section{Related Work}
In this section, we review related work on both classical and learning-based quadruped locomotion.

\subsection{Classical Quadruped Locomotion}
Quadruped locomotion has been a focus of the robotics research community for decades. Starting from building reliable legged hardware~\cite{raibert1986legged, raibert2008bigdog, hutter2016anymal, katz2019mini, di2020software}, researchers have also developed autonomous navigation and locomotion techniques to move these highly articulated robots in the real world~\cite{wooden2010autonomous, jeon2022online, sprowitz2013towards}. Most research into quadruped locomotion focus on adapting robot joint angles in the form of gaits~\cite{sprowitz2013towards} to challenging underlying terrain, including rugged rocks, non-rigid ground, and slippery surfaces. With sophisticated gait planners and controllers, quadruped robots can overcome the limitations of conventional wheeled mobile robots and successfully move through a variety of hard-to-reach spaces. However, the hyper redundancy of the many DoFs has led to capabilities beyond just overcoming the terrain challenges from underneath the robot: by carefully coordinating the limb joints, quadruped robots can also fit in confined 3D spaces through a combination of dexterous locomotion, manipulation, and navigation. 

One line of research into navigating legged robots through confined 3D spaces is based on whole-body planning-based methods. Mathieu et al.~\cite{geisert2019contact} showed that the acyclic gait planner, first introduced by Tonneau et al.~\cite{tonneau2018efficient}, can be applied to multilegged robots. Inspired by this work, Buchanan et al.~\cite{buchanan2021perceptive} integrated the acyclic gait planners into a hierarchical system that uses a high-level perceptive trajectory planner that plans intermediate body poses for the low-level gait planner to follow. Although the method improved upon Tonneau et al.~\cite{tonneau2018efficient} in a way that takes the precise collision models of the legs into consideration, which enables navigation in very confined 3D spaces, the system still suffers from relatively slow speed caused by frequent replanning and the complicated hierarchical structure. In contrast, our proposed approach is highly efficient, leveraging a sampling-based high-level reactive planner and an RL-based low-level locomotion policy to quickly produce joint angle commands and move the robot through confined 3D spaces. 

\subsection{Learning-Based Quadruped Locomotion}

Researchers from the robot learning community have also investigated quadruped locomotion using data-driven approaches~\cite{shao2021learning, fu2021minimizing, tan2018sim, yang2020data, pan2020zero}. Thanks to the vast amount of simulated trial-and-error experiences gathered using RL, quadruped robots are able to learn superior locomotion skills~\cite{rudin2022advanced, margolis2022rapid}, speed~\cite{bellegarda2022robust, jin2022high}, stability~\cite{kumar2021rma, agarwal2023legged, yu2021visual, saggar2007autonomous}, and energy efficiency~\cite{margolis2022rapid, hwangbo2019learning} compared to the classical methods. 

However, previous research on RL-based locomotion controllers primarily focuses on learning a limited set of gait parameters such as leg velocity, step frequency, and step amplitude. Such predefined gait patterns might be inefficient when maneuvering robots through complex and confined environments. To overcome these limitations, Rudin et al. \cite{rudin2022advanced} proposed to learn a locomotion controller using end-to-end RL training on position-based locomotion tasks. Specifically, the robot is only rewarded positively for reaching a goal position, which allows the agent to select its own path and gait beyond the predefined gaits. However, it is unclear whether acyclic and asymmetrical gait patterns can emerge from such end-to-end learning pipeline  in order to react in real-time to obstacles in confined 3D spaces. Moreover, such an end-to-end training can be inefficient when the goal is far away from the robot. In contrast, our proposed approach aims at using RL to efficiently enable the full potential of highly articulated quadruped robots beyond only challenging terrain but also confined 3D spaces that require close and accurate coordination of all the joint angles to fit the robot through 360\textdegree-surrounding obstacles (Fig.~\ref{fig::go1}). 
\section{Approach}
\label{sec:approach}
In this paper, the objective is to learn a quadruped locomotion controller for goal-reaching tasks in confined 3D spaces with RL. Each task in this context is distinctly characterized by a goal location denoted as $(x_g, y_g) \in \mathbb{R}^2$, representing a 2D coordinate within the world's reference frame. The locomotion controller is represented by a neural network policy $\pi(a\mid o)$ that takes as input observations $o$ and gives as output joint position commands $a$. This section introduces three distinct approaches: end-to-end dexterity, hierarchical dexterity, and parameterized motor skills.

\subsection{Observation and Action Spaces}
The observation and action spaces are shared across the three approaches except for the definition of commands. Specifically, an observation, denoted as $o = (g, q, \dot{q}, h, c)$, constitutes a five-element tuple, each of which is elaborated as follows:
\paragraph{Gravity Vector ($g \in \mathbb{R}^3$)} This element characterizes the orientation of the gravity vector within the robot's body frame, and it is acquired through the use of an Inertial Measurement Unit (IMU).
\paragraph{ Proprioceptive States ($q \in \mathbb{R}^{12}$ and $\dot{q} \in \mathbb{R}^{12}$)} These elements contain 12-dimensional joint angles and joint velocities measured by motor encoders.
\paragraph{Height Field ($h \in \mathbb{R}^{220}$)} This element is measured following a methodology similar to Agarwal et al.~\cite{agarwal2023legged}. However, it extends beyond the measurement of just the floor height to address terrain challenges. Instead, it captures both floor and ceiling heights to capture all-around obstacles. Precisely, the height field $h$ is represented as two $10 \times 11$ matrices, encompassing measurements of floor and ceiling heights. These measurements are taken at predefined $10 \times 11$ scandots evenly distributed across a 1-meter by 1-meter area situated in front of the robot. Subsequently, the height field is flattened and concatenated with the other components of the observation.
\paragraph{Commands (c)} Locomotion controllers operate under specified commands, which are included as a part of the observation. The definition of commands $c$ varies across the three different approaches. For example, the end-to-end dexterity and hierarchical dexterity are commanded with the global and local goal locations respectively. We provide the formal definitions of commands in the subsequent sections.

With regards to actions, each action $a \in \mathbb{R}^{12}$ assigns joint position targets at a frequency of 50 Hz for a Proportional-Derivative (PD) controller. The controller itself operates at a frequency of 200 Hz with the proportional and derivative gains remaining consistent with the values established in prior work~\cite{margolis2022rapid}.

\subsection{End-to-End Dexterity}
In the end-to-end dexterity approach, the commands denoted by $c$ are defined as the relative goal position $(x_r, y_r) \in \mathbb{R}^2$, representing the x-y coordinates within the robot's body frame of reference. This means that the policy always takes the relative position of the global goal, and it is trained to directly compute the low-level motor commands necessary for navigating to that global goal. Motivated by a similar end-to-end training pipeline introduced by Rudin et al.~\cite{rudin2022advanced}, the end-to-end dexterity approach incorporates four distinct reward terms: (1) goal-reaching reward; (2) stalling reward; (3) exploration reward; (4) penalty reward. The reward terms are elaborated as follows:
\paragraph{Goal-reaching reward}
\begin{equation}
r_g = \left\{
    \begin{array}{cc}
        +1, & \mbox{if } \sqrt{x_r^2 + y_r^2} < d_{g},  \\
        0, & \mbox{otherwise.}
    \end{array}
\right.
\end{equation}
This reward term assigns +1 when the agent reaching the global goal within a threshold distance denoted as $d_g$. Then, the episode terminateds. 
\paragraph{Stalling reward}
\begin{equation}
r_s = \left\{
    \begin{array}{cc}
        -1, & \mbox{if } \sqrt{v_x^2 + v_y^2} < 0.1 m/s,  \\
        0, & \mbox{otherwise.}
    \end{array}
\right.
\end{equation}
Here $v_x$ and $v_y$ are the robot's linear velocity in the x and y directions in robot's own frame of reference. The stalling reward penalizes the robot for standing still. A negative reward is assigned at every time step when the robot's linear velocity is smaller then 0.1 m/s. 
\paragraph{Exploration reward}
Due to the sparse nature of the main reward, it is beneficial to add a reward term encouraging exploration at the beginning of training. In order to bias the policy to walk towards the global goal, the exploration reward term incentivizes any base velocity in the correct direction. The reward is
defined as
$$
r_e = \frac{v_x x_r + v_yy_r}{\sqrt{(v_x^2 + v_y^2)(x_r^2 + y_r^2})}.
$$
\paragraph{Penalty reward}
We penalize joint accelerations, joint torques, collisions, and abrupt actions changes. The penalty reward is defined as follows
$$
r_{\mbox{penalties}} = - c_1||\ddot{q}||^2 - c_2 ||\tau||^2 - c_3 N_c - c_4 ||a - a_{-1}||^2.
$$
where $q$, $\tau$, $N_c$, $a$, $a_{-1}$ represent joint positions, joint torques, number of collisions, actions, and the actions from last time step respectively, and $c_{1-4}$ are scaling constants. A collision is recorded whenever the thighs, calf, or torso of the body collide with an obstacle or the floor. 

\subsection{Hierarchical Dexterity}
In contrast to direct navigation to the final goal position, the hierarchical dexterity approach involves a hierarchical system with a classical planner that constantly computes local goals, following which the robot can reach the global goal location, and a locomotion controller that receives commands of local goal locations. Therefore, $c$ is defined as the local goal position $g_l = (x_l, y_l)$ within the robot's body frame of reference.

However, planning in 3D space is very expensive and infeasible during real-world deployment where the entire 3D space is not known a priori, i.e., obstacles in the confined 3D space may obstruct the sensors' fields of view. Therefore, we employ a reactive sampling-based planner, which only uses the available information locally, checks the validity\footnote{To avoid expensive whole-body collision checking, the validity of a candidate pose is checked based on collisions between only the torso and the obstacles. Therefore, the planner only provides a rough guidance to the RL locomotion controller.} of a few pre-defined candidate poses, and selects the valid candidate pose that is closest to the final goal position. These candidate poses are expressed as 6-DoF states $p = (x, y, h_z, \phi, \theta, \psi)$, where $(x, y)$ represents a 2D coordinate in the body-frame, $h_z$ denotes the body height, and $(\phi, \theta, \psi)$ denote roll, pitch, and yaw angles, respectively. Then, the x-y coordinate of the selected candidate pose is used as the local goal location $(x_l, y_l)$. Notably, this planning process can be efficiently computed in parallel on a GPU.  

The reward design closely matches that of the end-to-end dexterity approach, with the global goal replaced by the local goal in all reward terms.

\subsection{Parameterized Motor Skills}
Instead of allowing the robot to learn its own motor skills, the parameterized motor skills approach only searches within the parameter space of a predefined motor skill. This approach operates under the assumption that these predefined motor skills are sufficiently effective for moving the robot through 3D confined spaces, and learning on this smaller solution space is potentially more efficient.

In this approach, the commands $c$ are defined as the parameters of motor skills, specifically denoted as $(v_x^{cmd}, v_y^{cmd}, \omega_z^{cmd}, h_z^{cmd}, \theta^{cmd}, \psi^{cmd})$. Here, $v_x^{cmd}$ and $v_y^{cmd}$ represent the target linear velocities in the body-frame x- and y-axes, while $\omega_z^{cmd}$ signifies the target angular velocity around the yaw axis. Additionally, $h_z^{cmd}$, $\theta^{cmd}$, and $\psi^{cmd}$ correspond to the target height, roll, and pitch angles, respectively.

Notably, unlike prior works that mostly train the policy on randomly sampled commands~\cite{margolis2022rapid, margolis2022walktheseways}, the parameterized motor skills approach trains the policy on the real command distribution of the task of moving through 3D confined spaces. To achieve this, the approach utilizes the same reactive sampling-based planner employed in the hierarchical dexterity approach. During training, the commands always direct the robot toward the planned local goal poses. Specifically, let $p_l = (x_l, y_l, h_{zl}, \phi_l, \theta_l, \psi_l)$ be the body pose of the planned target local goal. The target linear velocity is a constant 0.5m/s, directed toward the local goal location, which is defined by $v_x^{cmd} = 0.5 \cdot x_l / \sqrt{x_l^2 + y_l^2}$ and $v_y^{cmd} = 0.5 \cdot y_l / \sqrt{x_l^2 + y_l^2}$. $\omega_z^{cmd}$ is a constant 1.57 rad/s, maintaining a fixed rate of rotation towards the target yaw angle. $h_z^{cmd}$, $\theta^{cmd}$, and $\psi^{cmd}$ are kept the same as the target local goal.

To learn these parameterized motor skills, the approach incorporates a \textit{velocity tracking reward} that encourages the robot's linear and angular velocity to be the same as the target velocities, and a \textit{pose tracking reward} that penalizes deviations of the robot's height, roll, and pitch angles from the target orientations. The rewards are formally defined as follows.
\paragraph{Velocity Tracking Reward} This reward component motivates the agent to reach the local goal with a constant linear velocity pointing to the local goal location and a constant angular velocity turning the robot to the target yaw angle. The velocity tracking reward contains two reward terms $r_{vxy}$ and $r_{wz}$ defined as
\begin{equation}
   r_{vxy} = \exp\{\frac{-(v_x - v_x^{cmd})^2 - (v_y - v_y^{cmd})^2}{\sigma_{vxy}}\} 
\end{equation}
\begin{equation}
    r_{wz} = \exp\{\frac{-(w_z - w_z^{cmd})^2}{\sigma_{wz}}\}  
\end{equation}
Here $w_z$ is the robot's angular velocity around the yaw axis. $\sigma_{vxy}$ and $\sigma_{wz}$ are two hyper-parameters.
\paragraph{Pose Tracking Reward} This reward component penalizes the robot for not using the target body poses to reach the local goal position. It contains the following three reward terms: 
\begin{align*}
    r_{h} = (h_z - h_z^{cmd}), r_\theta = (\theta - \theta^{cmd}), r_{\psi} = (\psi - \psi^{cmd}).
\end{align*}

\section{Experiments}
\label{sec:expriment}
This section provides an overview of the training processes for the three approaches specified in Section~\ref{sec:approach}. In Section~\ref{sec:env}, we describe the procedural pipeline used for generating confined 3D environments, while Section~\ref{sec:training} delves into the details of training and evaluating the learned locomotion policies for each approach.

\subsection{Environments}
\label{sec:env}
\begin{figure}
    \centering
    \includegraphics[width=0.9\columnwidth]{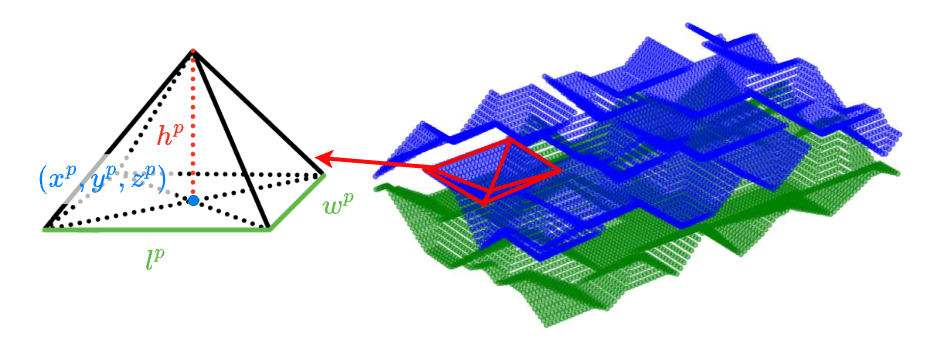}
    \caption{An example of confined 3D environments constructed from random pyramids (right) and an illustration of pyramid parameters (left).}
    \label{fig:pyramid}
\end{figure}
The 3D confined environments are constructed from randomly generated $2 \times n_r \times n_c$ matrices of pyramids, positioned on both the ceilings and the floors. As shown in Fig.~\ref{fig:pyramid}, each pyramid is uniquely characterized by a a set of parameters $(x^p, y^p, z^p, l^p, w^p, h^p)$. Here, $x^p$, $y^p$, and $z^p$ represent the 3D coordinate of the center of the base. The base itself is a rectangle with $l^p$ and $w^p$ denoting its length and width, while $h^p$ represents the height of the tip.

The set of parameters of an element indexed by $(u, i, j)$ from the $2 \times n_r \times n_c$ pyramid matrix is denoted as $(x^p_{u,i,j}, y^p_{u,i,j}, z^p_{u,i,j}, l^p_{u,i,j}, w^p_{u,i,j}, h^p_{u,i,j})$, signifying a pyramid located at row $i$ and column $j$, with $u$ taking values from the set $\{0, 1\}$ to distinguish between pyramids on the floor ($u=0$) and those on the ceiling ($u=1$). The range of values for these parameters are specified in Table~\ref{tab:pyramid_param}. Here, $\mbox{Uniform}(a, b)$ denotes a uniform distribution in the range of $[a, b]$.
\begin{table}[htb!]
\centering
\begin{tabular}{cc}
\\
\toprule
Parameter & Values \\
\midrule
$x^p_{u,i,j}$      &    $\Delta x \cdot i + \mbox{{Uniform}}(-0.3, 0.3)$    \\
$y^p_{u,i,j}$      &    $\Delta y \cdot j + \mbox{Uniform}(-0.3, 0.3)$   \\
$z^p_{u,i,j}$      &    $0.5 \cdot u$   \\
$l^p_{u,i,j}$      &    $\mbox{Uniform}(0.2, 0.4)$    \\
$w^p_{u,i,j}$      &    $\mbox{Uniform}(0.2, 0.4)$    \\
$h^p_{u,i,j}$      &    $(2u - 1) \cdot \mbox{{Uniform}}(0.15, 0.35) $ \\
\bottomrule
\end{tabular}
\caption{Pyramid parameters}
\label{tab:pyramid_param}
\end{table}

The x-y coordinates of the pyramids roughly adhere to a 2D grid with slight variances randomly sampled from the range of $[-0.3, 0.3]$. In the table, $\Delta x$ and $\Delta y$ are intervals of the grid in x- and y-axis. The height of the base is set to 0 for pyramids on the floor and 0.5 meters for pyramids on the ceiling. The length and width of the base are randomly sampled from the range of $[0.2, 0.4]$. The height of the tip is randomly sampled from the range of $[0.15, 0.35]$. Notably, the height of the tip is negative for pyramids on the ceiling to reflect their inverted orientation.

\begin{figure}
\vspace{-10pt}
    \centering
    \includegraphics[width=0.9\columnwidth]{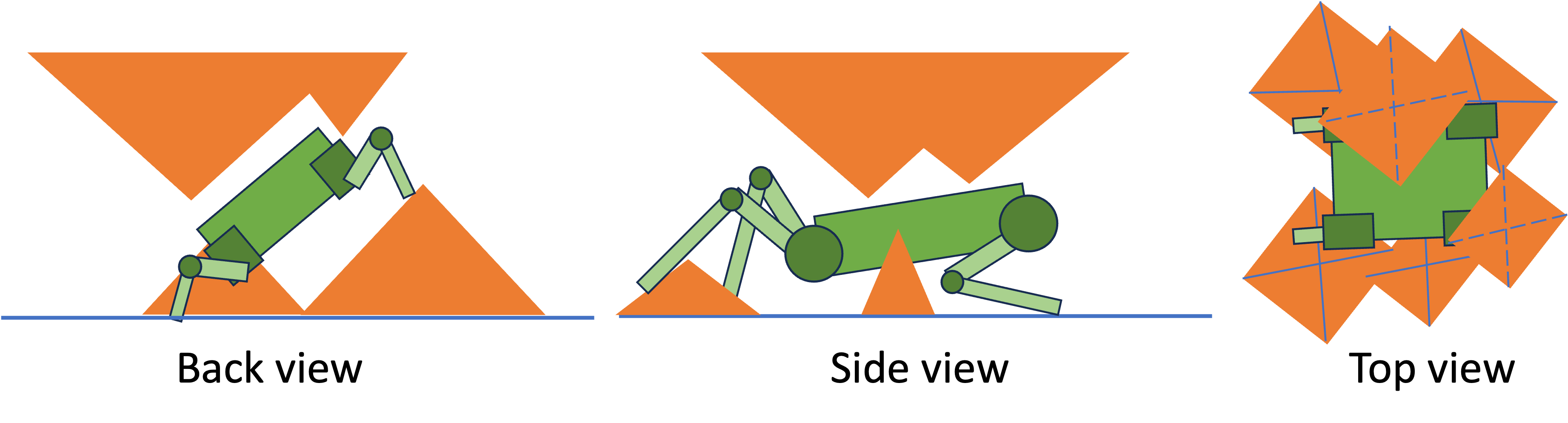}
    \caption{Different views of a legged robot moving through confined 3D environments composed of random pyramids.}
    \label{fig:robot_in_tunnel}
\end{figure}

These 3D environments are categorized into three difficulty levels: \emph{easy}, \emph{medium}, and \emph{hard}, with the number of columns $n_c$ set to 2, 3, and 4, respectively. The number of rows, denoted as $n_r$, remains constant at 3. The size of the pyramid matrices remains the same for all three difficulty levels to cover an area of 2.5 meters by 1.5 meters. Consequently, environments with more columns or rows include denser pyramids, introducing greater challenges for traversal. To navigate these 3D obstacles, the robots are initialized at the point $(-1.75, 0)$, positioning itself 0.5 meters away from the pyramids. The navigation task is to reach the goal location at $(+1.75, 0)$, situated 0.5 meters beyond the pyramids. Fig.~\ref{fig:robot_in_tunnel} shows different views of a robot navigating in a 3D confined environment.

\subsection{Training Details}
\label{sec:training}

\begin{figure}
    \centering
    \includegraphics[width=0.65\columnwidth]{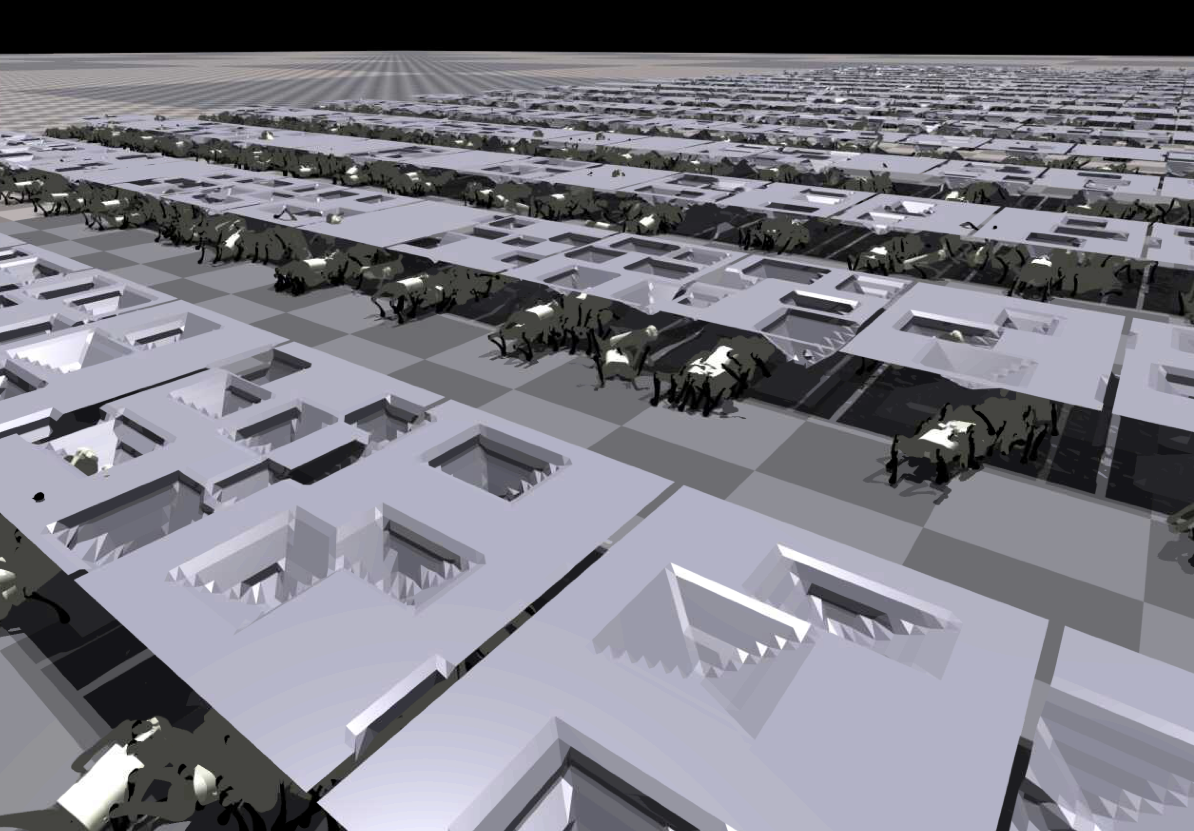}
    \caption{Massive parallel training in Isaac Gym.}
    \label{fig:isaac}
\end{figure}

The implementation of the three approaches is based on the massive parallel training pipeline introduced by Margolis and Agrawal \cite{margolis2022walktheseways}, employing the Isaac Gym~\cite{narang2022factory} simulation and the Proximal Policy Optimization (PPO) algorithm~\cite{schulman2017proximal} as the core Deep RL technique. The policies are trained in parallel to control a Unitree Go1 robot to navigate on 1000 \emph{easy}, 1500 \emph{medium}, and 1500 \emph{hard} environments describe in Section \ref{sec:env}. Fig~\ref{fig:isaac} visualizes the massive parallel training in Isaac Gym.

\paragraph{Domain randomization for Sim-to-real transfer} For better sim-to-real transfer, we randomize a wide range of parameters during training, including the robot’s body mass, motor strength, joint position calibration, ground friction and restitution, and orientation and magnitude of gravity.

\paragraph{Curriculum}Due to the sparse reward used by the end-to-end dexterity approach, we employ a curriculum strategy. This strategy initiates training with an initial x-coordinate goal location, set at $x_g = 0.6$. Then, the value is incremented by $0.2$ meters whenever the success rate of reaching the current goal locations surpasses a threshold of $40\%$. Importantly, this curriculum strategy is specifically applied to the end-to-end dexterity approach and is not employed for the other two approaches.
\section{Results}
The trained policies are tested on both simulated and real Go1 robots. This section describes the testing results from these two settings.
\subsection{Simulated results}
The trained policies are evaluated on 100 \emph{easy}, 100 \emph{medium}, and 100 \emph{hard} environments. The evaluation runs 10 independent trials in the simulation for each environment, resulting in a total of 3000 test trials. 
The approximate training steps required for convergence, average success rates during testing, and average collision counts of all successful test trials are reported in Tables \ref{tab:efficienty}, \ref{tab:success_rate}, and \ref{tab:collision} respectively. We provide in-depth analysis as follows. 

\begin{table}
\vspace{-10pt}
\centering
\begin{adjustbox}{width=\columnwidth,center}
\begin{tabular}{ccc}
\\ \toprule
End-to-end dexterity               & Hierarchical dexterity & Parameterized motor skills \\
\midrule
600 million  &  \textbf{40 million}    &   80 million     \\
\bottomrule
\end{tabular}
\end{adjustbox}
\caption{Number of time steps to convergence}
\label{tab:efficienty}
\vspace{-10pt}
\end{table}

\begin{table}
\centering
\begin{tabular}{cccc}
\\ \toprule
Approach                 & Easy & Medium & Hard \\
\midrule
End-to-end dexterity &   29.4\%   &   7.0\%     & 5.1\%      \\
Hierarchical dexterity (ours) &   \textbf{96.8\%}   &    \textbf{51.5\%}    &   \textbf{39.4\%}   \\
Parameterized motor skills             &   65.7\%   &  4.0\%     & 2.4\%    \\
\bottomrule
\end{tabular}
\caption{Success rates of the three approaches on confined 3D environments with three difficulty levels.}
\label{tab:success_rate}
\end{table}

\begin{table}
\vspace{-10pt}
\centering
\begin{tabular}{cccc}
\\ \toprule
Approach                 & Easy & Medium & Hard \\
\midrule
End-to-end dexterity &   7.9   &   \textbf{16.2}     &  \textbf{29.8}      \\
Hierarchical dexterity &   \textbf{3.8}  &   32.3    &   43.6   \\
\bottomrule
\end{tabular}
\caption{Average Collision Count. Bold numbers indicate better collision avoidance.}
\label{tab:collision}
\end{table}

\subsubsection{Learning Efficiency}
As shown in Table \ref{tab:efficienty}, during training, the hierarchical dexterity approach demonstrates significantly superior learning efficiency. It converges after approximately 40 million environment steps, whereas the end-to-end dexterity approach requires roughly 600 million environment steps. The simulation runs at a speed of about 5000 steps per second on a single-GPU machine, which attributes to 2.2 hours and 33.3 hours of training for hierarchical dexterity and  end-to-end dexterity respectively.

We attribute the enhanced performance of the hierarchical dexterity approach to its distinct separation of long-term navigation and short-term motor skill acquisition, in contrast to the end-to-end version. 
To be specific, the end-to-end training approach attempts to encompass the entire solution space for dexterous locomotion in confined 3D spaces. However, such an approach is shown to be highly inefficient in our experiments, particularly when dealing with sparse rewards. In contrast, the hierarchical dexterity approach preserves the solution space to effectively explore motor skills but at the same time only needs to reach short-term local goals computed by a classical planner. Such a hierarchical approach provides denser reward signals and therefore facilitates more efficient learning of the motor skills which are essential for successful traversal within confined 3D spaces.

Acquiring predefined motor skills takes approximately 80 million environment steps, which is inferior but comparable in terms of sample efficiency to the hierarchical dexterity approach. However, these predefined motor skills do not enable sufficient dexterity in confined 3D spaces.

\subsubsection{Learned Dexterity}
As shown in Table \ref{tab:success_rate}, the hierarchical dexterity approach outperforms the other two approaches for all three difficulty levels and consistently maintain a significant advantage. In \emph{easy} environments, the hierarchical dexterity approach almost finishes every single trial, while the parameterized motor skills approach achieves more than twice the success rate of end-to-end learning. It is likely that the limited solution space provided by the parameterized motor skills is sufficient to go through sparse 3D obstacles. 
Challenges arise when transitioning to more demanding environments, characterized as \emph{medium} or \emph{hard}, which necessitate advanced motor skills beyond those predefined ones. In such scenarios, the success rate experiences a significant decline, underscoring the need for end-to-end or hierarchical training to preserve the full solution space for an agent to explore its own motor skills.
The success rates of the end-to-end approach and parameterized motor skills approach both drop to single digits for \emph{medium} and lower single digits for \emph{hard}. Although performance also drops for the hierarchical dexterity approach, it can still finish half of the environments in \emph{medium} and more than one third in \emph{hard}. 

\subsubsection{Obstacle-Aided Locomotion}
We also show the average collision count measured during all successful trials of the end-to-end and hierarchical dexterity approaches in Table \ref{tab:collision}, which reveals an interesting finding in terms of obstacle-aided locomotion. While the collision count for end-to-end dexterity increases with more difficult environments, it does not change much for hierarchical dexterity. Overall, hierarchical dexterity has higher collision count compared to end-to-end learning. Such an observation shows that hierarchical dexterity is more willing to touch obstacles and still achieves a higher success rate than its more conservative counterpart. Considering the all-around obstacle constraints in confined 3D spaces, it is likely that the robot has to lean against some obstacles to assure torso stability and forward progress. Guided by a local goal close to the robot, it can focus more on discovering these emergent motor skills, e.g., obstacle-aided locomotion despite the small collision penalty, to utilize the environment structure to move, rather than trying to figure out how to navigate to a global goal far away from the robot. 

\subsection{Real-World Demonstration}

We demonstrate the policy learned in simulation on a physical Unitree Go1 quadruped robot by creating an artificial setup replicating the simulated environments (Fig.~\ref{fig::go1}). This setup allows us to assess the policy's performance in a controlled real-world scenario. The robot generates a point cloud from its front-facing stereo camera, which is used to build real-time height maps during deployment. We filter out the points to limit the robot's observation to cover only a 1m$\times$1m$\times$0.5m space directly in front of it. Height maps of the ceiling and floor are then generated from the filtered points based on the pitch and roll angle of the camera detected by an onboard IMU and passed to the policy. To enhance the quality of the height maps and mitigate potential noise resulting from the stereo camera's point cloud generation process, we apply some basic smoothing. 
In this experiment, we configure the goal location to always be one meter in front of the robot to ensure continuous forward motion.
Our real-world demonstration shows that the robot can navigate through confined 3D spaces, while occasionally touching the obstacles made from rocks and cardboard boxes, using the controller learned in simulation.

\section{Conclusions and Future Work}
This paper introduces a novel and challenging scenario in the realm of legged locomotion. In this scenario, legged robots are tasked with traversing confined 3D spaces that impose constraints from all around the robot, demanding the adaptation of acyclic and asymmetric locomotion behaviors. To tackle this challenging scenario, three distinct approaches are proposed in this paper: end-to-end dexterity, hybrid dexterity, and predefined motor skills. Empirical results point to the hybrid dexterity approach as the most effective and promising solution for locomotion in confined 3D spaces. This approach combines high-level pose planning from a classical planner with low-level local goal-reaching via an RL-based locomotion controller.

While the experiments conducted in this paper utilize randomly generated pyramid environments, it's important to acknowledge that these environments are artificial and may not accurately represent real-world 3D confined spaces, such as tunnels or other complex structures. In the future, exploring training on real-world tunnels or constructing more realistic simulation environments to capture a wider variety of confined 3D spaces remains an important direction.

Furthermore, the paper's focus on a simple reactive planner is noteworthy. This planner may not be able to adequately address scenarios with highly complex navigation paths. An open question remains whether it is feasible to employ more advanced path planning techniques specifically designed for 3D spaces, while ensuring compatibility with the massive training pipeline for legged locomotion in IsaacGym. 

\newpage
\bibliographystyle{IEEEtran}
\bibliography{IEEEabrv,references}

\end{document}